\pdfoutput=1

\documentclass[letterpaper, 10 pt, conference]{ieeeconf}  % Comment this line out if you need a4paper

\IEEEoverridecommandlockouts                              % This command is only needed if 
\usepackage{amsmath}
\usepackage{graphicx}
\usepackage{amssymb}
\usepackage{booktabs}
\usepackage{multirow}
\usepackage{multicol}
\usepackage{float}
\usepackage{subfigure}
\usepackage{kotex}
\usepackage{cite}
\usepackage{tabularx}
\usepackage{array}
\usepackage[flushleft]{threeparttable}
\usepackage[ruled,vlined]{algorithm2e}
\usepackage{bm}

\begin{document}
\title{\bf Avoiding Degeneracy for Monocular Visual SLAM \\ with Point and Line Features}

\author{Hyunjun Lim$^{1}$, Yeeun Kim$^{1}$, Kwangik Jung$^{2}$, Sumin Hu$^{1}$, and Hyun Myung$^{*}, $~\IEEEmembership{Senior~Member, IEEE}%
\thanks{$^{1}$H. Lim, $^{1}$Y. Kim, and $^{1}$S. Hu are with School of Electrical Engineering, Korea Advanced Institute of Science and Technology~(KAIST), Daejeon, Republic of Korea
{\tt\small \{tp02134, yeeunk, 2minus1\}@kaist.ac.kr}}%        
\thanks{$^{2}$K. Jung is with Department of Civil and Environmental Engineering of KAIST, Daejeon, Republic of Korea
{\tt\small ankh88324@kaist.ac.kr}}%
\thanks{$^{*}$H. Myung is with School of Electrical Engineering, KI-AI, and KI-R, KAIST, Daejeon, Republic of Korea
{\tt\small hmyung@kaist.ac.kr}}%
\thanks{This work was supported by the Defense Challengeable Future Technology Program of Agency for Defense Development, Republic of Korea. The students are supported by Korea Ministry of Land, Infrastructure and Transport (MOLIT) as “Innovative Talent Education Program for Smart City” and BK21 FOUR.}% <-this % stops a space
% \textit{(Corresponding author: Hyun Myung)}
}


\markboth{IEEE Robotics and Automation Letters. Preprint Version. Accepted Month, Year}
{FirstAuthorSurname \MakeLowercase{\textit{et al.}}: ShortTitle} 



\maketitle

\begin{abstract}
In this paper, a degeneracy avoidance method for a point and line based visual SLAM algorithm is proposed. Visual SLAM predominantly uses point features. However, point features lack robustness in low texture and illuminance variant environments. Therefore, line features are used to compensate the weaknesses of point features. In addition, point features are poor in representing discernable features for the naked eye, meaning mapped point features cannot be recognized. To overcome the limitations above, line features were actively employed in previous studies. However, since degeneracy arises in the process of using line features, this paper attempts to solve this problem. First, a simple method to identify degenerate lines is presented. In addition, a novel structural constraint is proposed to avoid the degeneracy problem. At last, a point and line based monocular SLAM system using a robust optical-flow based lien tracking method is implemented. The results are verified using experiments with the EuRoC dataset and compared with other state-of-the-art algorithms. It is proven that our method yields more accurate localization as well as mapping results. 
\end{abstract}

%\begin{IEEEkeywords}
%\end{IEEEkeywords}

% \fi
\IEEEpeerreviewmaketitle

\section{Introduction}
% for IEEE journal papers produced under \LaTeX\ using
% IEEEtran.cls version 1.8b and later. \\

As a result of the advancement of unmanned mobile robots, Simultaneous Localization and Mapping (SLAM) has been extensively studied. Because GPS typically does not function in indoor environments, SLAM based on cameras, LiDARs, and Inertial Measurement Units (IMUs) have become attractive research topics. Among these sensors, cameras are the most widely used ones as they have become low-cost and easily accessible\cite{scaramuzza2019visual, huang2019visual}. Thus, visual SLAM have emerged as an essential technology in robotics, augmented/virtual reality, and autonomous driving. 
% Recently, algorithms that fuse with other sensors have been developed to compensate for the weakness of the camera's low output rate and fragility against fast movements.

\begin{figure}[ht]
    \centering
    \includegraphics[width=1\linewidth]{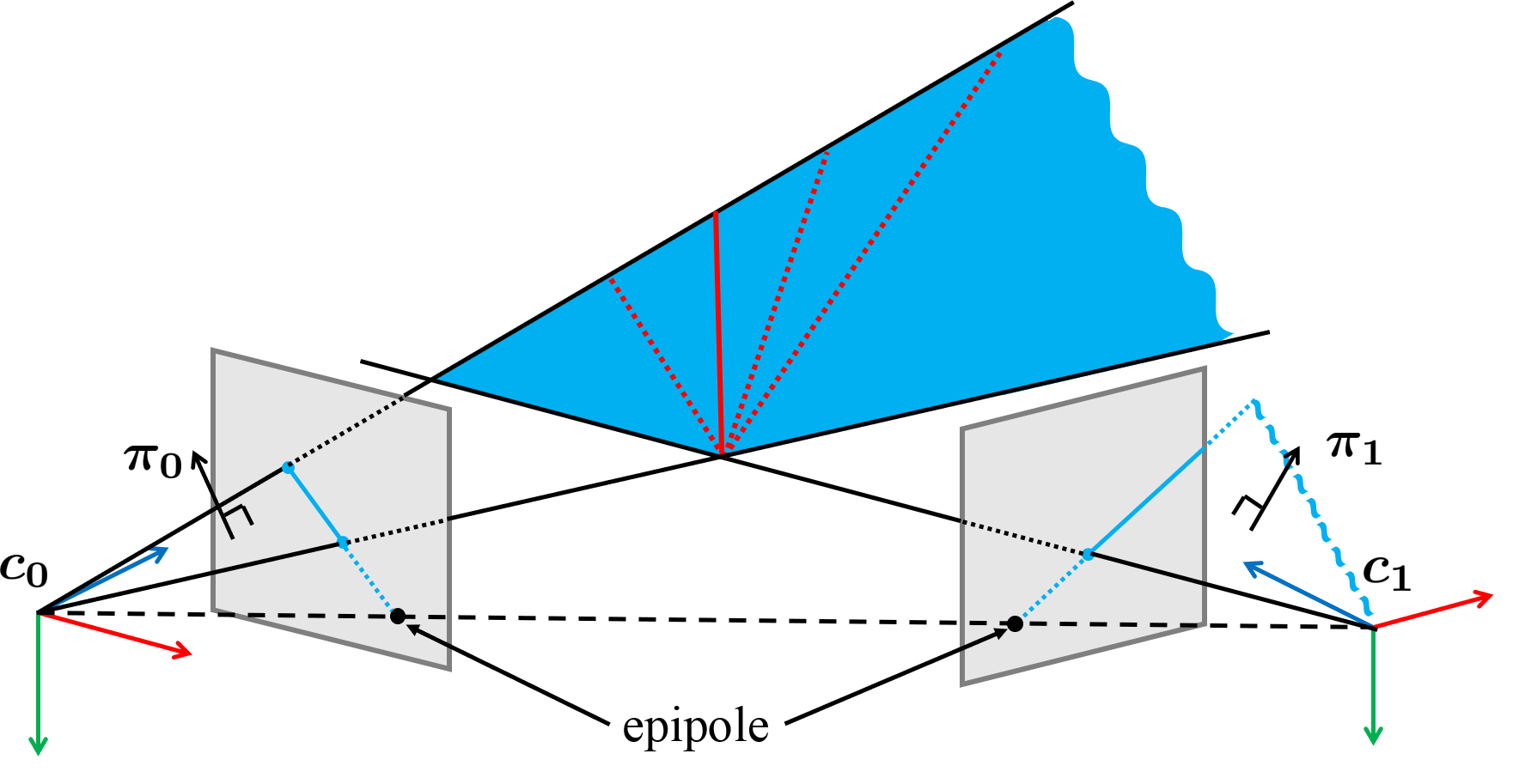}
    \caption{Illustration of degeneracy. Degeneracy occurs when line is in proximity to the epipole. In the triangulation process, $\boldsymbol{\pi}_0$ and $\boldsymbol{\pi}_1$ formed by each camera center and the observed line feature, coincide with the epipolar plane. Then, the estimated 3D line should be on a plane colored in blue. Although the actual 3D line is a red solid line, triangulation can mis-determine the estimated line as one of the infinitely many dotted lines on the plane.
    }
    \label{fig:degeneracy}
\end{figure}

Nowadays, Visual SLAM algorithms are predominantly feature-based method due to its fast performance. However, since most feature-based algorithms are based on point features, there is a high probability that descriptors may change and tracking fails in situations with large illumination variances or low-texture \cite{kottas2013efficient, kong2015tightly}. In addition, as most point-based algorithms are sparse, the identification of the environment structure in the mapping result is challenging.

Since such shortcomings inherently exist for point features, some studies have shown that line features are better for visual SLAM. A line feature consists of multiple points, making it resilient against environmental changes. Also, it serves as an excellent landmark even in environments with low texture such as corridors. Furthermore, mapping with line features reveal the geometry of the surrounding, becoming easily discernible to the human eye. 

However, under degeneracy conditions, line features are no longer robust. Degeneracy refers to a phenomenon in which a line feature does not manifest as a single 3D line during triangulation. 
This occurs when the lines observed in both images pass through the epipoles. In this case, as shown in Fig.~\ref{fig:degeneracy}, the planes $\boldsymbol{\pi}_0$ and $\boldsymbol{\pi}_1$ formed by each camera center and the observed line feature, coincide with the epipolar plane. Then, the estimated 3D line should be on a plane colored in blue. Although the actual 3D line is a red solid line, triangulation can mis-determine the estimated line as one of the infinitely many dotted lines on the plane. As a result, the 3D line feature cannot be properly determined, resulting in large reconstruction errors. The degeneracy problem occurs more often when using line features than point features. This is because degeneracy of a point feature occurs only when the point is close to the epipole, whereas in the case of a line feature, the distances of all points included in the line feature must be distant from the epipole. As degeneracy affects the localization accuracy and the mapping result, it must be addressed in the line feature-based SLAM.

In this paper, we propose a variation of visual SLAM that can avoids the degeneracy problem of line features stated above. The main contributions of this study are as follows:
\begin{itemize}
    \item A simple method to identify degenerate lines in pure translation is presented.
    \item A novel structural constraint is proposed to avoid the degeneracy problem.
    \item A point-and-line-based monocular SLAM system using a robust optical-flow-based line tracking method is proposed.
    % \item Using point and line features, an optimization-based monocular SLAM is proposed running in real time.
    % \item A robust optical-flow-based line tracking method is proposed to solve the failure that occurs during the line tracking process.
\end{itemize}

The remainder of this paper is organized as follows. Section~\ref{sec:related work} provides an overview of related works. Section~\ref{sec:proposed method} describes the proposed method in detail. Section \ref{sec:experimental results} presents the experimental results. Finally, Section \ref{sec:conclusion} summarizes our contributions and discusses future work.
%\IEEEpubidadjcol

\section{Related works}\label{sec:related work}
% In general, feature-based SLAM tracks only point features in an image. After that, the cameras' pose and the 3D position of the features are corrected through bundle adjustment using the point feature's re-projection error as a cost function. 
Until recently, methods using point features were predominant in SLAM\cite{klein2009parallel, mur2015orb} and Visual-Inertial System (VINS)\cite{jones2011visual, bloesch2015robust, mourikis2007multi, leutenegger2015keyframe, qin2018vins} algorithms. The reasons were clear. Point features can be extracted from any image. In addition, the comparison process is simple because it can be expressed in 2D coordinate on the image. However, point features have poor tracking performance in environments with low texture or illumination variance. Furthermore, the sparsity of the estimated 3D point features makes them inappropriate for mapping. To improve the performance, some studies focused on line features instead.

Among the many studies with line features, Bartoli \textit{et al.} \cite{bartoli2005structure} laid the foundations on how to use line features in SLAM. They proposed a method using the line feature in structure-from-motion. Since line features have a higher degree of freedom than point features, a parameterization method was required. They used Pl\"{u}cker coordinates and orthonormal representations. Also, the re-projection error of the line feature was proposed for bundle adjustment. Zhang \textit{et al.} \cite{zhang2015building} went further by implementing a filtering-based stereo SLAM with line features using the method above.

Through the studies above, existing point feature-based algorithm have been extended to incorporate line features. First, Zheng \textit{et al.} \cite{zheng2018trifo} and Yang \textit{et al.} \cite{8967905} used line features based on MSCKF\cite{mourikis2007multi}. Next, various algorithms\cite {pumarola2017pl, gomez2019pl, zuo2017robust, lee2019elaborate} that add line features based on ORB-SLAM\cite{mur2015orb} have been proposed. Lastly, He \textit{et al.} \cite{he2018pl} and Fu \textit{et al.} \cite{fu2020pl} applied line feature based on VINS-Mono\cite{qin2018vins}. Through the studies above, it was possible to improve localization performance. In addition, structural characteristics of the surrounding environment could be obtained through the mapping of line features. 

However, degenerate lines appeared in line-based SLAM. The commonly used re-projection error was vulnerable to degeneracy. A line is defined by its normal and direction vectors in the Pl\"{u}cker coordinates\cite{bartoli2005structure}. However, the reprojection error is calculated only by a line’s normal vector, meaning that the optimization process can only affect this vector. In degenerate cases, lines have anomalous direction vectors which cannot be corrected with the reprojection error alone.

The degeneracy of the line feature is much more severe than that of the point feature \cite{hartley2003multiple}. Therefore, there were studies \cite{sugiura20153d, zhou2018slam} that analyzed the degeneracy occurring in the process of applying line features to visual SLAM. In addition, the MSCKF-based approach proposed by Yang \textit{et al.} \cite{8967905} analyzed the degenerate motions in two different line triangulation methods. In such studies, when line degeneracy occurs, all the corresponding lines were removed. Therefore, the loss of features was a limitation to the method. As a method to solve the degeneracy of lines, Ok \textit{et al.} \cite{ok2012accurate} proposed to reconstruct 3D line segment with artificial points if the image lines are nearly-aligned with the epipolar line. However, this method has limited applications in that it can only be employed when degenerate lines are in contact with other lines.

\begin{figure*}[ht!]
    \centering
    \subfigure[]{\label{fig:translation1}\includegraphics[width=0.49\linewidth]{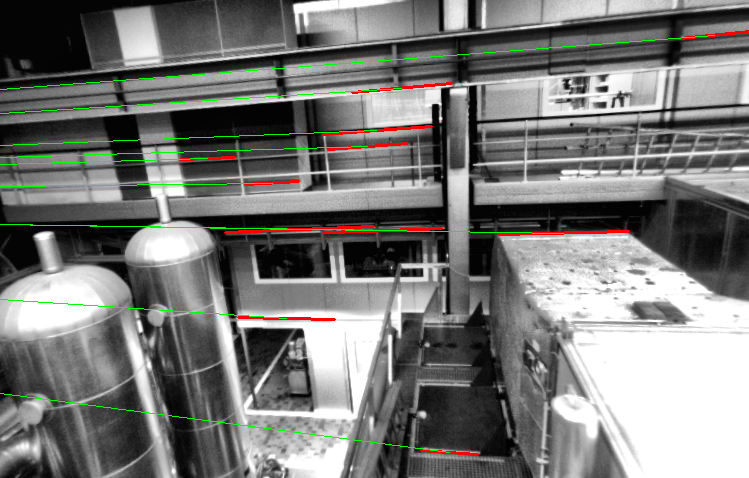}}
    \subfigure[]{\label{fig:translation2}\includegraphics[width=0.49\linewidth]{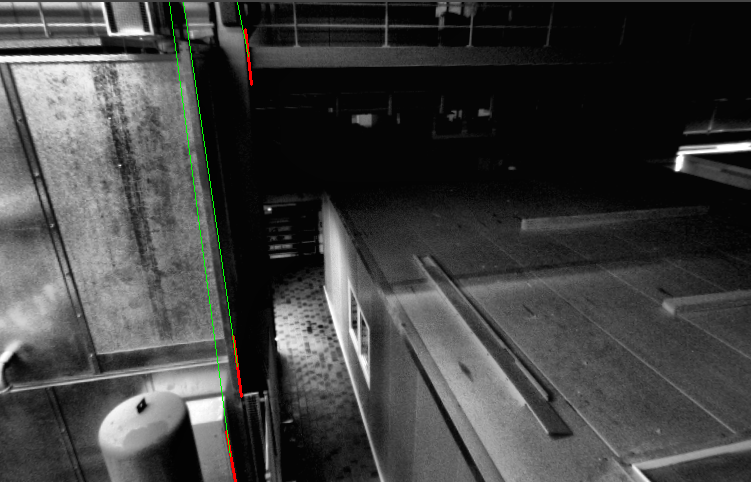}}
    \caption{Extracted lines of the image where degeneracy occurs in the pure translation of a camera in the EuRoC dataset\cite{burri2016euroc}. The red lines are the extracted lines, and the green lines are the lines connecting the epipole and midpoint of the extracted line. \subref{fig:translation1}~The camera moves along the $x$-axis, and degeneracy occurs in all lines along the $x$-axis direction. \subref{fig:translation2}~The camera moves along the $y$-axis, and degeneracy occurs in all lines along the $y$-axis direction.}
    \label{fig:translation}
\end{figure*}

\section{Preliminaries}
A point feature in 3D can be intuitively represented as $(x, y, z)$. In contrast, a line feature has 4-DoF, which can be represented in numerous manners \cite{bartoli2005structure}. In this study, we use two different methods of representation: Pl$\ddot{\text{u}}$cker coordinates and orthonormal representation. 

\subsection{Pl$\ddot{\text{u}}$cker Coordinates}
Pl$\ddot{\text{u}}$cker coordinates are represented by 6-DoF consisting of a direction and a normal vector, as shown in Fig.~{\ref{fig:line:plucker}}. It can intuitively represent line features and is advantageous in obtaining the 3D line by triangulation or projecting the line to a 2D plane. With the 3D direction vector as $\boldsymbol{d}$ and the normal vector as $\boldsymbol{n}$, the mathematical representation of a 3D line is as follows:
\begin{equation} 
\boldsymbol{L}({\boldsymbol{n}}^\top, {\boldsymbol{d}}^\top)^\top \in \mathbb{R}^6.\\
\label{eq:plucker coordinates}
\end{equation}

\subsection{Orthonormal Representation}
Orthonormal representation is a minimal 4-DoF representation of a 3D line. It does not suffer from over-parameterization as Pl$\ddot{\text{u}}$cker coordinates do during the optimization step. The 3D line’s orthonormal representation is as follows:
\begin{equation}
\boldsymbol{o}_j = [\psi_1, \psi_2, \psi_3, \phi]
\label{eq:orthonormal representation}
\end{equation}
where $\psi_j$ is the $j$-th column of the 3D line's rotation matrix with respect to the camera coordinate system in Euler angles, and $\phi$ is the parameter of minimal distance from the camera center to a point on the line.

To optimize the 3D line features, we convert the over-parameterized Pl$\ddot{\text{u}}$cker coordinates representation to a minimal orthonormal representation. First, the Pl$\ddot{\text{u}}$cker coordinates are decomposed by QR decomposition \cite{bartoli2005structure}, as follows: 
\begin{equation}
\begin{bmatrix}
{\boldsymbol{n}} & {\boldsymbol{d}}
\end{bmatrix}=
\begin{bmatrix}
{\boldsymbol{n}} \over \parallel{\boldsymbol{n}}\parallel & 
{\boldsymbol{d}} \over \parallel{\boldsymbol{d}}\parallel &
{\boldsymbol{n}} \times {\boldsymbol{d}} \over \parallel {\boldsymbol{n}} \times {\boldsymbol{d}}\parallel\\
\end{bmatrix}
\begin{bmatrix}
\parallel{\boldsymbol{n}}\parallel & 0 \\
0 & \parallel{\boldsymbol{d}}\parallel \\
0 & 0
\end{bmatrix}.
\label{eq:QR}
\end{equation}
The first component on the right side in \eqref{eq:QR} represents the rotation matrix $\boldsymbol{U}$, which is a function of the Euler angle $\boldsymbol{\psi}$, as follows:
\begin{equation}
\boldsymbol{U} = \boldsymbol{R}(\boldsymbol{\psi})=
\begin{bmatrix}
{\boldsymbol{n}} \over \parallel{\boldsymbol{n}}\parallel & 
{\boldsymbol{d}} \over \parallel{\boldsymbol{d}}\parallel &
{\boldsymbol{n}} \times {\boldsymbol{d}} \over \parallel {\boldsymbol{n}} \times {\boldsymbol{d}}\parallel\\
\end{bmatrix}.
\label{eq:rotation matrix U}
\end{equation}

In addition, $\boldsymbol{W}$ calculated from normal and direction vector is used to calculate the 1-DoF $\phi$. It contains information about the minimal distance, $\parallel{\boldsymbol{d}}\parallel \over \parallel{\boldsymbol{n}}\parallel$, between the camera coordinate and a 3D line feature, as follows: 
\begin{equation}
\small
\begin{split}
\boldsymbol{W} =&
\begin{bmatrix}
\cos\phi & -\sin\phi \\
\sin\phi & \cos\phi
\end{bmatrix} \\ 
=&
{1\over\sqrt{{\parallel {\boldsymbol{n}} \parallel}^2 + {\parallel {\boldsymbol{d}} \parallel}^2}}
\begin{bmatrix}
\parallel {\boldsymbol{n}} \parallel & -\parallel {\boldsymbol{d}} \parallel\\
\parallel {\boldsymbol{d}} \parallel & \parallel {\boldsymbol{n}} \parallel
\end{bmatrix}.    
\end{split}
\label{eq:W}
\end{equation}

\section{The Proposed Method}\label{sec:proposed method}
The proposed algorithm is an extension of VINS-Mono to line features that overcome degeneracy problems. Because this algorithm has roots in VINS-Mono, IMU measurements use the pre-integration method \cite{forster2016manifold}; optimization uses two-way marginalization with Schur complement \cite{sibley2010sliding}; and point feature manipulation uses optical-flow, triangulation, and re-projection error. The additional part of our work consists of Section~\MakeUppercase{\romannumeral 3}.$\textit{A}$, degeneracy analysis, Section~\MakeUppercase{\romannumeral 3}.$\textit{B}$, degeneracy avoidance using structural constraint, and Section~\MakeUppercase{\romannumeral 3}.$\textit{C}$, line-based SLAM.

\subsection{Degeneracy Identification}
\label{sec:degeneracy}

In general, degeneracy can be determined by calculating the distance to the epipole. To determine degeneracy, the position of the epipole must be calculated during the triangulation process. The equation for the position of the epipole is as follows:
\begin{equation}
    \boldsymbol{p}_{epipole} = {\boldsymbol{R}^w_{c_j}}^\top(\boldsymbol{t}^w_{c_i} - \boldsymbol{t}^w_{c_j})
    \label{eq:epipole}
\end{equation}
where $\boldsymbol{p}_{epipole}$ is the epipole of the $i$-th camera coordinate $c_i$ with respect to the $j$-th camera coordinate $c_j$. $\boldsymbol{p}_{epipole}$ is calculated by rotation matrix $\boldsymbol{R}^w_{c_j}$ and translation $\boldsymbol{t}^w_{c}$ of camera with respect to world coordinate. As shown in \eqref{eq:epipole}, the position of the epipole is affected by the camera motion. 
  
Particularly in pure translation, degeneracy conditions for line features are more susceptable. If the camera moves in pure translation, the epipole is located at infinity in the direction of the camera's movement. In this situation, when 2D lines are considered straight lines rather than line segments, all lines prallel to camera's movement can be considered to pass through the epipole. For example, in Fig.~{\ref{fig:translation}} \subref{fig:translation1} and \subref{fig:translation2}, the camera's movement is on the $x$ and $y$-axis directions, respectively. In this paper, when the camera moves in pure translation, line features parallel to the camera motion are determined which is when degeneracy occurs.

\subsection{Degeneracy Avoidance Using Structural Constraint}
\label{sec:structural constraint}

When degeneracy of a line feature occurs, an erroneous direction vector is obtained during the mapping process. Through the line triangulation process \eqref{eq:make_plane}, the direction vector of the line, $\boldsymbol{d}$, is obtained as follows:
\begin{equation}
\boldsymbol{d} = 
\begin{bmatrix}
\pi_{0z}\pi_{1y} - \pi_{0y}\pi_{1z} \\
\pi_{0x}\pi_{1z} - \pi_{0z}\pi_{1x} \\
\pi_{0y}\pi_{1x} - \pi_{0x}\pi_{1y} \\
\end{bmatrix}. 
\label{eq:degeneracy_triangulation}
\end{equation}
When the epipole is close to the line feature, the two plane vectors, $\boldsymbol{\pi}_0(\pi_{0x}, \pi_{0y}, \pi_{0z })$ and $\boldsymbol{\pi}_1(\pi_{1x}, \pi_{1y}, \pi_{1z})$ are parallel as shown in Fig.~{\ref{fig:degeneracy}}. Therefore, in theory, with each element of $\boldsymbol{d}$ in \eqref{eq:degeneracy_triangulation} as 0, $\boldsymbol{d}$ becomes a zero matrix. However, because the measurement noise of this line feature might exist, an erroneous direction vector might be calculated. Hence, when degeneracy occurs in this way, the direction vector of Pl$\ddot{\text{u}}$cker coordinates are affected and anomalous. Finally, A poor mapping result is obtained due to degenerate lines. 

However, the line's re-projection error in \eqref{eq:line_projection} mainly used in the optimization process cannot solve the degeneracy. Because line's re-projectoin error uses only the normal vector, it cannot correct the direction vector's error. Existing algorithms resolved the degeneracy problem by removing degenerate lines. However, this approach leads to a loss of valuable line feature information. Therefore, there is a need to correct the line direction vector without removing the degenerate line features.

\begin{figure}[t]
    \centering
    \includegraphics[width=0.9\linewidth]{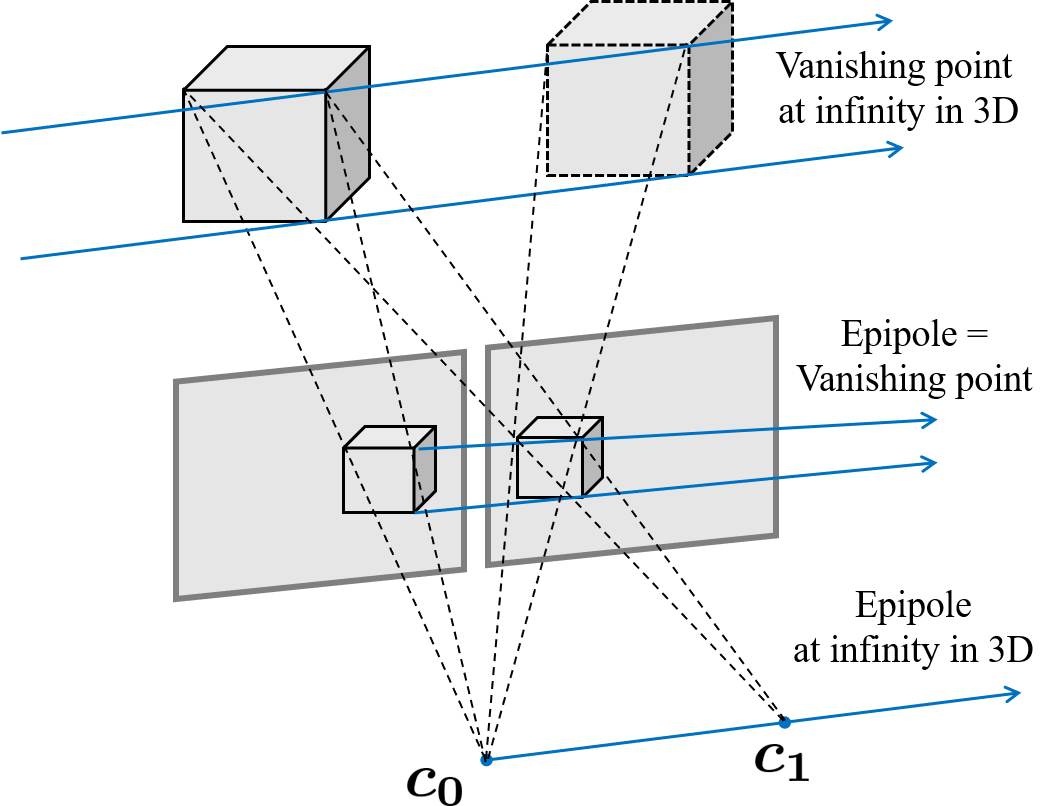}
    \caption{Illustration showing that the vanishing point and the epipole are the same in the pure translation along $x$-axis. The object indicated by the dotted line is the translation of the camera expressed as its movement. As straight lines passing through the same point of an object are parallel, they have a common vanishing point at infinity in 3D. In addition, as it is the pure translation, the epipole is located at infinity in 3D. Therefore, the vanishing point and the epipole are identical when projected onto the 2D image.}
    \label{fig:vp_and_ep}
    % \vspace{-5mm}
\end{figure}

Through the degeneracy analysis, it was found that degeneracy often occurs in camera motion of pure translation. In addition, we consider the fact that the epipole and the vanishing point have the same position in pure translational camera motion \cite{hartley2003multiple} as shown in Fig.~{\ref{fig:vp_and_ep}}. Thus, degenerate lines always have vanishing points in the baseline direction. Therefore, lines with the same vanishing point can be corrected for lines with degeneracy using the property that they are parallel in 3D. We define this as a structural constraint, and the residual for the parallel condition is:
\begin{equation}
\mathbf{r}_{\mathcal{S}}(\mathbf{z}_{ij}, \mathcal{X})=
\boldsymbol{d}_{i} \times \boldsymbol{d}_{j}
\label{eq:structrual_constraint}
\end{equation} 
where $\boldsymbol{d}_i$ and $\boldsymbol{d}_j$ denote the direction vectors of the $i$-th and $j$-th lines, respectively. Thus, because a constraint is created by pairing all lines where degeneracy occurs, the accuracy of mapping can be improved by creating parallel lines. Compared to other methods \cite{zhou2015structslam, zou2019structvio}, such as using a J-linkage\cite{toldo2008robust, tardif2009non} to find parallel lines to the vanishing point, this method is more efficient

\begin{figure*}[ht!]
    \centering
    \subfigure[]{\label{fig:line:plucker}\includegraphics[width=0.285\linewidth]{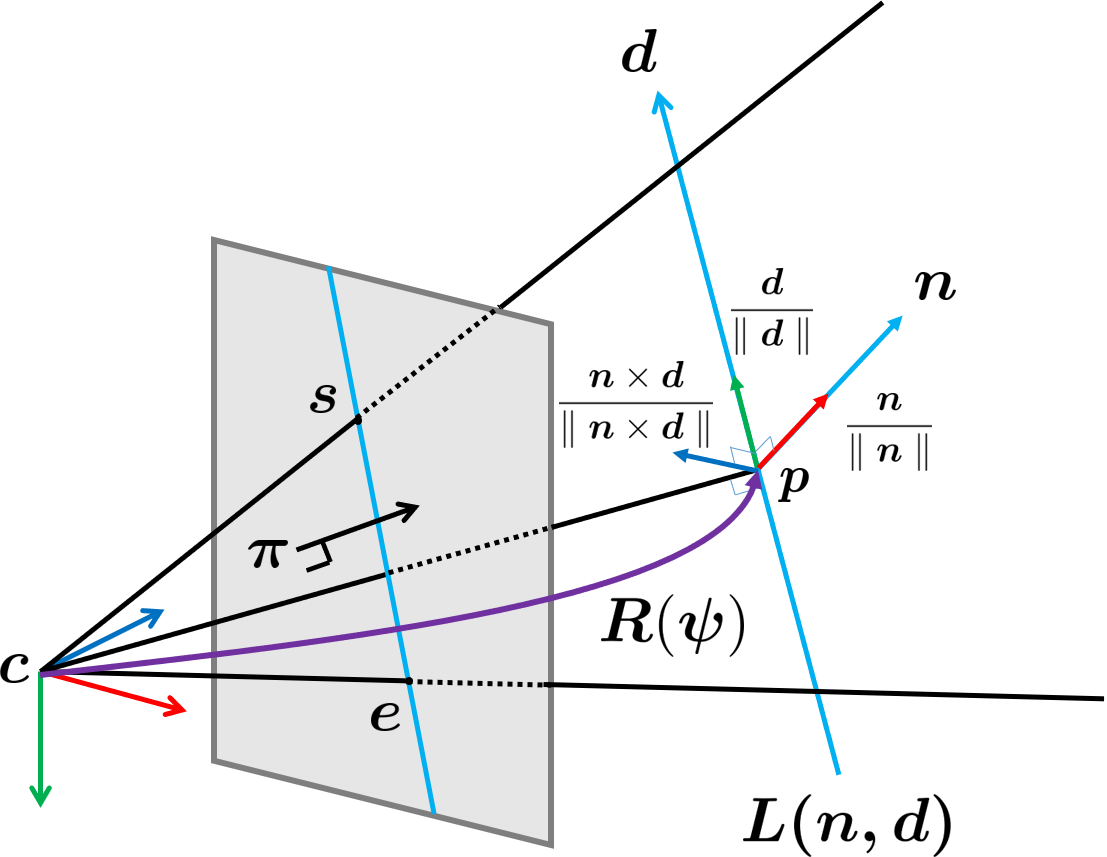}}
    \subfigure[]{\label{fig:line:triangulation}\includegraphics[width=0.365\linewidth]{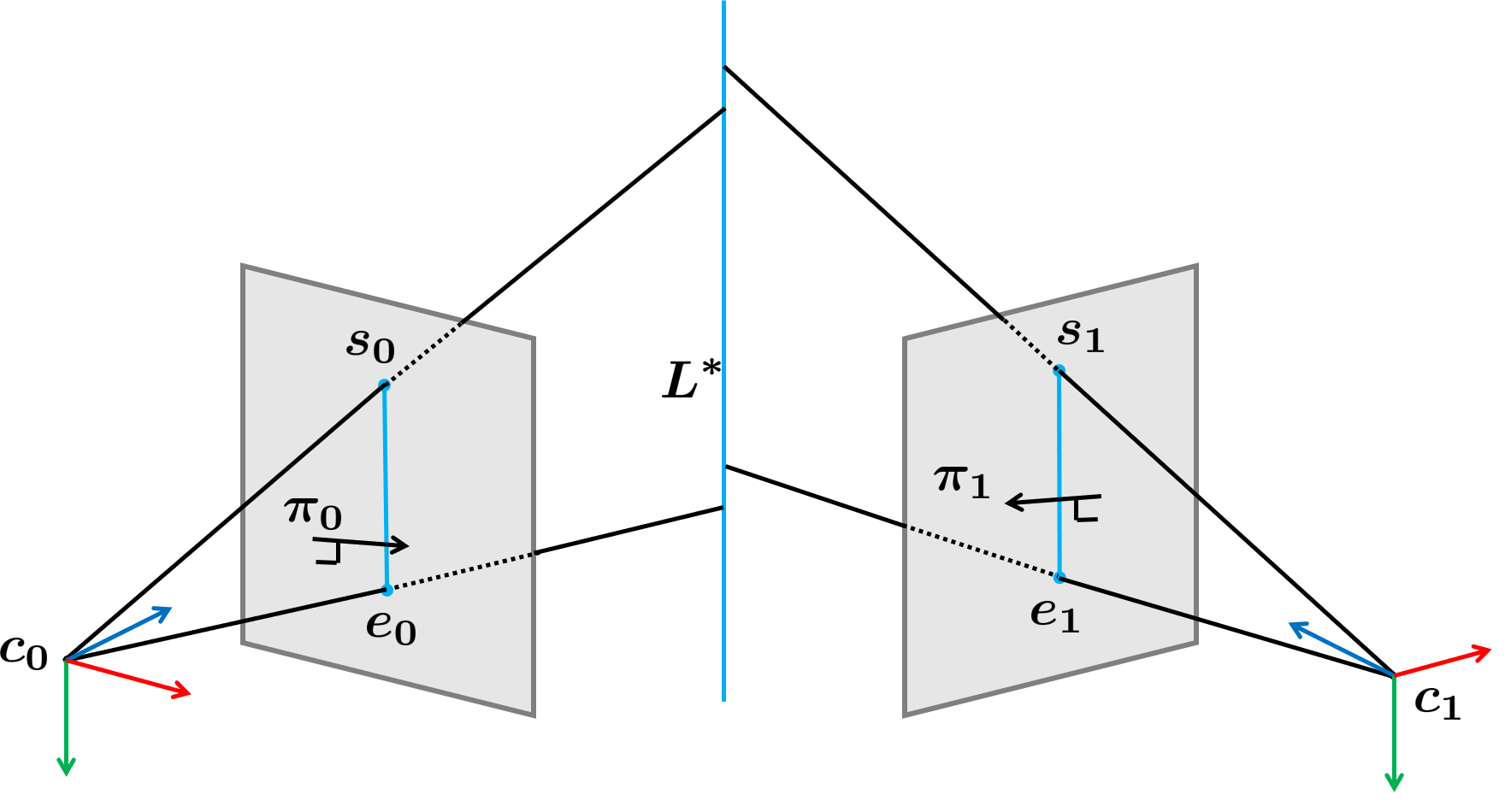}}
    \subfigure[]{\label{fig:line:projection}\includegraphics[width=0.325\linewidth]{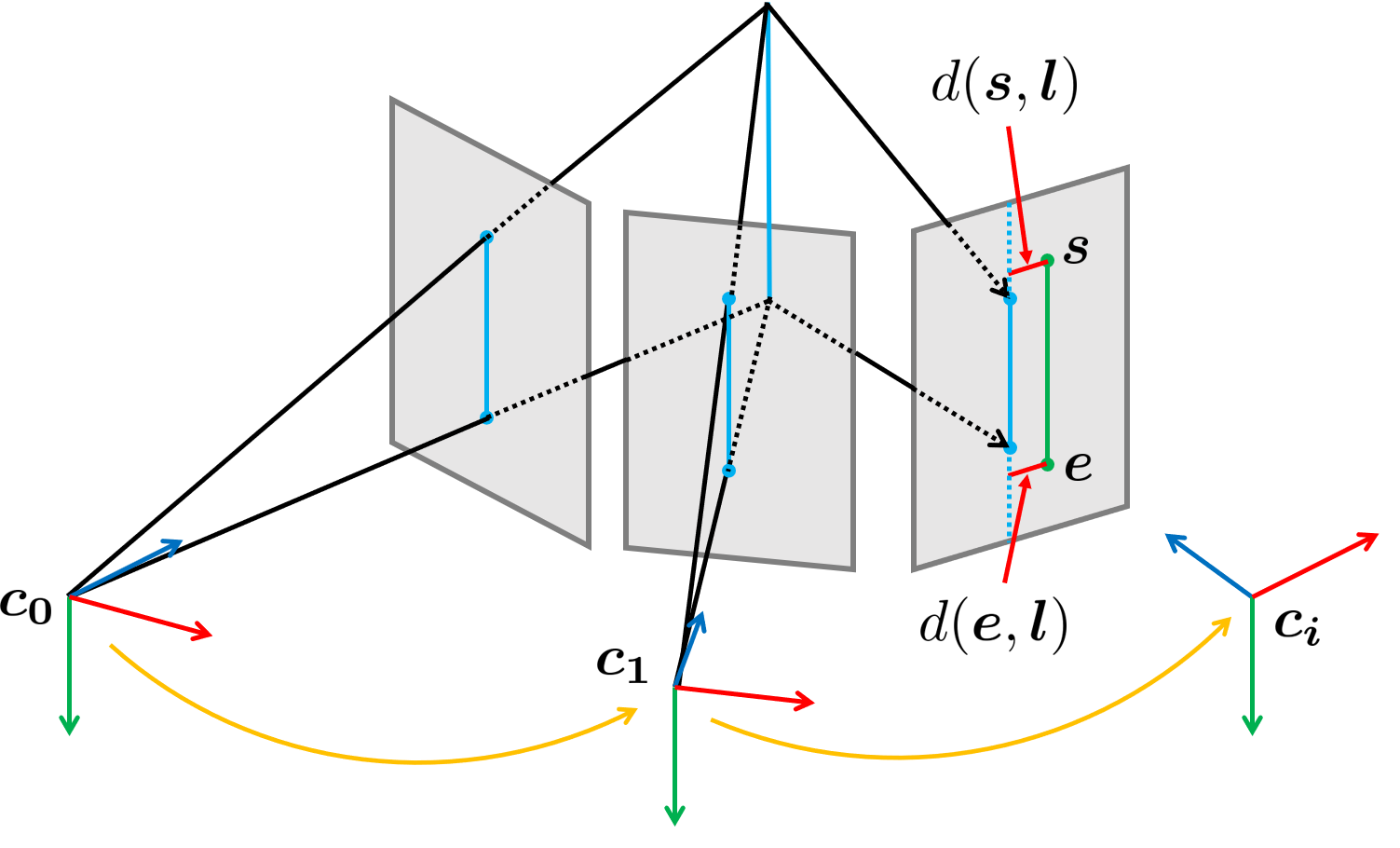}}
    \caption{Illustration of line-based SLAM. \subref{fig:line:plucker} Pl\"{u}cker coordinates consist of normal and direction vectors. In addition, orthonormal representation consists of Euler angles of the line's rotation matrix and parameter of distance. \subref{fig:line:triangulation} Each plane can be obtained through the camera position and line endpoints. The intersection of two planes creates a 3D line. \subref{fig:line:projection} The re-projection of the line is used to define the cost function used in the optimization process. The distance between the re-projected line endpoints and the observed line is used as a residual.}
    \label{fig:line}
\end{figure*}

\begin{figure*}[ht]
    \centering
    \includegraphics[width=1\linewidth]{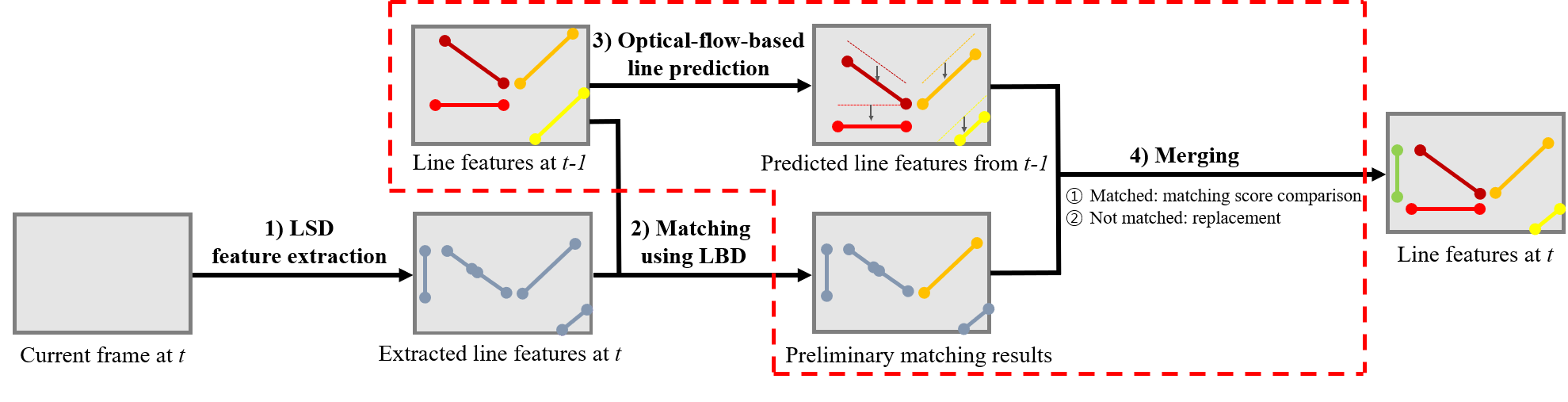}
    \caption{Flowchart of line tracking. Between adjacent frames, the same lines are represented with the same color. By using LSD and LBD, lines are extracted and matched, respectively. In addition, the predicted lines are made by optical-flow. By comparing the predicted lines and the lines created through matching, it is possible to remove ther cut or duplicate lines.}
    \label{fig:line_tracking}
\end{figure*}

\subsection{Line-based SLAM}
\label{sec:line-based SLAM}

\subsubsection{State Definition}
The state vector used in our system is as follows:
\begin{equation}
\begin{aligned}
\mathcal{X} = [&\boldsymbol{x}_{0}, \boldsymbol{x}_{1}, \cdots ,\boldsymbol{x}_{k-1},\\
&\lambda_{0}, \lambda_{1}, \cdots, \lambda_{m-1},\\ 
&\boldsymbol{o}_{0}, \boldsymbol{o}_{1}, \cdots, \boldsymbol{o}_{n-1}],\\
\boldsymbol{x}_i = [&\boldsymbol{p}^{w}_{b_i}, \boldsymbol{q}^{w}_{b_i}, \boldsymbol{v}^{w}_{b_i}, \boldsymbol{b}_{a}, \boldsymbol{b}_{g}],\\
\boldsymbol{o}_j = [&\boldsymbol{\psi}_j, \phi_j]
\label{eq:state}
\end{aligned}
\end{equation}
where $\mathcal{X}$ is the whole state and $\boldsymbol{x}_i$ is the state in the $i$-th sliding window and is composed of position, quaternion, velocity, and biases of the accelerometer and gyroscope in the appearing order. In addition, the whole state includes the inverse depths of point features represented as $\lambda$. Moreover, orthonormal representations of the lines, $\boldsymbol{o}$, which are newly added, are included.

\subsubsection{Triangulation}
When two lines from two different frames are matched, a 3D line feature can be estimated using triangulation, as shown in Fig.~{\ref{fig:line:triangulation}}. A 2D line feature and a camera coordinate system can define a 3D plane as follows:
\begin{align}
\begin{gathered}
\boldsymbol{\pi}_i = [\pi_{ix}, \pi_{iy}, \pi_{iz}, \pi_{iw}]^\top,\\
\begin{bmatrix}
\pi_{ix}\\\pi_{iy}\\\pi_{iz} 
\end{bmatrix} 
= [\boldsymbol{s}_i]_\times \boldsymbol{e}_i,\\ \pi_{ix} c_{ix} + \pi_{iy} c_{iy} + \pi_{iz} c_{iz} + \pi_{iw} = 0\\
\end{gathered}
\label{eq:make_plane}
\end{align}
where $\boldsymbol{\pi}_i$ is the $i$-th plane vector in sliding windows. $\boldsymbol{\pi}_i$ consists of values ($\pi_{ix}$, $\pi_{iy}$, $\pi_{iz}$) that can be calculated using the cross product of $\boldsymbol{s}_i$ and $\boldsymbol{e}_i$, which are the two endpoints of a detected line segment, and $\pi_{iw}$ that can be calculated using the camera center, $\boldsymbol{c}_i=(c_{ix}, c_{iy}, c_{iz}$). The two planes created by two camera coordinate systems, $\boldsymbol{c}_0$ and $\boldsymbol{c}_1$, intersect at a single 3D line, which becomes the 3D line feature. The dual Pl$\ddot{\text{u}}$cker matrix to determine a line intersecting two planes can be represented as follows:
\begin{equation}
\boldsymbol{L}^* = 
\begin{bmatrix}
[\boldsymbol{d}]_\times & {\boldsymbol{n}}\\
-{\boldsymbol{n}}^\top & 0
\end{bmatrix} = 
\boldsymbol{\pi}_0\boldsymbol{\pi}_1^\top-\boldsymbol{\pi}_1\boldsymbol{\pi}_0^\top.
\label{eq:dual_plucker}
\end{equation}

\subsubsection{Measurement Model}
The Pl$\ddot{\text{u}}$cker coordinates are transformed into an orthonormal representation by optimally finding the value that minimizes the cost function. For this reason, the 2D projection to an image plane should be obtained using the intrinsic parameter, as follows:
\begin{equation}
\boldsymbol{l} = 
\begin{bmatrix}l_1 \\ l_2 \\ l_3\end{bmatrix}
= \boldsymbol{K}{\boldsymbol{n}}^c =
\begin{bmatrix}
f_y & 0 & 0 \\
0 & f_x & 0 \\
-f_yc_x & -f_xc_y & f_xf_y
\end{bmatrix}
{\boldsymbol{n}}^c
\label{eq:line_projection}
\end{equation}
where $\boldsymbol{l}$ represents the re-projected line, $\boldsymbol{K}$ represents the camera's intrinsic parameter, and $\boldsymbol{n}^c$ represents re-projected line's normal vector. As this study uses normalized images, the intrinsic parameter is an identity matrix. Therefore, the normal of the 3D line feature is equal to the re-projected line.

In the case of Fig.~{\ref{fig:line:projection}}, the projected 3D line feature does not have an end point. Therefore, the cost function, or the residual is calculated by finding the distance between the detected 2D line feature endpoints and the 3D line feature, as follow:
\begin{align}
\begin{gathered}
\mathbf{r}_\mathcal{L}(\mathbf{z}_l^c, \mathcal{X})=
\begin{bmatrix}
d(\boldsymbol{s}^{c}, \boldsymbol{l}^{c}) \\
d(\boldsymbol{e}^{c}, \boldsymbol{l}^{c})
\end{bmatrix},\\
d(\boldsymbol{p},\boldsymbol{l})=
{\boldsymbol{p}^\top\boldsymbol{l} \over {\sqrt{l_1^2+l_2^2}}}
\end{gathered}
\label{eq:line residual}
\end{align}
where $\boldsymbol{r}_\mathcal{L}$ is the residual of the line feature and $\mathbf{z}_l^c$ is the measurement of detected line $\boldsymbol{l}$ with respect to camera coordinate $c$. And $\boldsymbol{r}_\mathcal{L}$ consists of the distance between $\boldsymbol{l}$ and endpoint $\boldsymbol{p}$ ($\boldsymbol{s}^c$ or $\boldsymbol{e}^c$). 

Using the state defined in \eqref{eq:state}, the overall cost function for optimization is set as follows:
\begin{equation}
\begin{gathered}
\min_{\mathcal{X}} \left\{
\parallel \mathbf{r}_p - \mathbf{H}_p \mathcal{X} \parallel ^2 
+ \sum_{k\in \mathcal{B}}\parallel{\mathbf{r}_{\mathcal{B}}
(\hat{\mathbf{z}}^{b_k}_{b_{k+1}}, \mathcal{X}) \parallel }^2_{\mathbf{P}^{b_k}_{b_{k+1}}} 
\right. \\ \left.
+ \sum_{(i,j) \in \mathcal{P}} \parallel{\mathbf{r}_{\mathcal{P}}
(\hat{\mathbf{z}}^{c_i}_{p_j}, \mathcal{X}) \parallel }^2_{\mathbf{P}^{c_i}_{p_j}} 
+ \sum_{(i,j) \in \mathcal{L}} \parallel{\mathbf{r}_{\mathcal{L}}
(\hat{\mathbf{z}}^{c_i}_{l_j}, \mathcal{X}) \parallel }^2_{\mathbf{P}^{c_i}_{l_j}} 
\right. \\ \left.
+ \sum_{i \in \mathcal{L}^*}\sum_{j \in \mathcal{L}^*} \parallel{\mathbf{r}_{\mathcal{S}}
(\hat{\mathbf{z}}_{ij}, \mathcal{X}) \parallel }^2 
\right\} 
\label{eq:total residual}
\end{gathered}
\end{equation}
where $\mathbf{r}_p$, $\mathbf{r}_{\mathcal{B}}$, and $\mathbf{r}_{\mathcal{P}}$ represent marginalization, IMU, and point measurement factor, respectively. These are the same with VINS-Mono. The newly added factors in the cost function are $\mathbf{r}_{\mathcal{L}}$ and $\mathbf{r}_{\mathcal{S}}$, which represent line measurement and structural constraints, respectively. In this case, $\mathcal{L}^*$ means parallel lines. In the optimization process, Ceres Solver\cite{ceres-solver} was used.

\subsubsection{Optical-flow-based Line Tracking}
\label{sec:optical-flow-based line tracking}
To utilize line features in SLAM, it is necessary to extract the lines and match them with lines extracted from the previous frame. Conventional SLAM algorithms, adopting line features, employ a Line Segment Detector (LSD) \cite{von2008lsd,von2012lsd} for their line extraction process. Line Band Descriptor (LBD) \cite{zhang2013efficient} was applied to match the extracted line features with the previous ones. In the previous studies \cite{he2018pl, gomez2019pl, 8967905}, LSD and LBD were used to filter out incorrectly matched lines through information such as orientation, length, and endpoint positions. However, there are limitations to this method. First, a single line can be split into multiple lines as the frame progresses. In addition, lines at the boundary of the image plane can be cut off. And those lines are considered as different lines and it causes tracking performance degradation. 

In this study, we propose a method based on optical-flow \cite{lucas1981iterative} to robustly track lines for a long time. The overall procedure can be seen in Fig.~{\ref{fig:line_tracking}}. When the current frame at $t$ is given, 1) first, line features are extracted using LSD. In this process, the short line features are removed for robustness. 2) Extracted line features are matched with the previous line features at the  ($t-1$)-th frame. But, as seen in the preliminary matching results, divided, or cut off lines are not matched, even if they should be considered as the same lines. To solve this problem, we add one more step: 3) optical-flow-based line prediction. In this step, the new line features at $t$ are predicted by transferring the lines’ endpoints observed at $t-1$ through optical-flow. And it is checked whether the predicted line features from $t-1$ are misidentified lines through matching with the line features at $t$. Then, these predicted line features are used to supplement the preliminary matching results. Among the line features in the preliminary matching results, there can exist the lines that are not matched but should have been matched. In case of matched lines, the matching scores between the extracted lines and their corresponding lines from optical-flow are compared, and the line features which have higher score are selected for final line features at $t$. When the lines are not matched, extracted line features are replaced with the corresponding line features from optical-flow. These processes correspond to the 4) merging step. Therefore, more robust line features can be utilized, and the line features can be continuously tracked more robustly.

\section{Experimental Results}\label{sec:experimental results}
The experiments were conducted with an Intel Core i7-9700K CPU with 32GB of memory. We evaluated the proposed algorithm using the EuRoC Micro Aerial Vehicle (MAV) dataset \cite{burri2016euroc}. Each dataset has its own level of difficulty depending on factors such as illumination, texture, and MAV speed. Therefore, it was suitable to test the effectiveness of the proposed method. TABLE~\ref{table:parameter} summarizes the parameters used in the experiments.
\begin{table}[H] 
\centering
\caption{Parameter settings in experiments}
\label{table:parameter}
\begin{tabular}{@{}cc@{}}
\toprule
Parameter & Value \\ \midrule
Sliding window size & 10 \\
Maximum number of point features & 150 \\
Minimum pixel length of line feature & 50 \\
Minimum number of tracked frames of line feature & 5 \\
Maximum solver time for optimization (s) & 0.1 \\
Maximum iterations for optimization & 10 \\ \bottomrule
\end{tabular}
\end{table}

To evaluate the performance of line tracking, we exploited only the lines tracked for a long time to reduce the computational cost in the optimization process. Among all lines, the ratio of the number of lines observed in sliding windows of a certain number of frames or more was calculated. In this experiment, only the lines observed for at least five sliding windows were used. As shown in TABLE~\ref{table:tracking result}, the optical-flow-based line tracking has approximately 49\% higher tracking performance than the conventional method with LSD and LBD while the computation time was increased by about 20\%.

\begin{table}[h]
\centering
\caption{tracking performance of line features with the EuRoC dataset. Percentage (\%) of the number of lines tracked in 5 or more sliding windows.}
\label{table:tracking result}
\begin{tabular}{ccc}
\toprule
                  & LSD+LBD (\%) & Proposed (\%)       \\
\midrule                  
MH\_01\_easy      & 23.48   & \textbf{49.76} \\
MH\_02\_easy      & 28.35   & \textbf{51.30} \\
MH\_03\_medium    & 25.24   & \textbf{43.35} \\
MH\_04\_difficult & 30.58   & \textbf{51.92} \\
MH\_05\_difficult & 31.46   & \textbf{53.82} \\
V1\_01\_easy      & 34.51   & \textbf{45.21} \\
V1\_02\_medium    & 24.95   & \textbf{33.43} \\
V1\_03\_difficult & 19.96   & \textbf{27.32} \\
V2\_01\_easy      & 37.56   & \textbf{47.59} \\
V2\_02\_medium    & 24.15   & \textbf{28.69} \\
V2\_03\_difficult & 19.16   & \textbf{22.60} \\
\bottomrule
\end{tabular}
% \vspace{-5mm}
\end{table}

The accuracy of the proposed algorithm is compared with the accuracy of VINS-Mono without loop-closing. In addition, to prove the efficacy of degeneracy avoidance, we compare the proposed algorithm with PL-VINS\cite{fu2020pl} which is a state-of-the-art algorithm that uses line features and the code has been released.  

\begin{figure*}[ht]
    \centering
    \begin{minipage}[c]{0.51\textwidth}
    \subfigure[]{\label{fig:mapping:a}\includegraphics[width=\linewidth]{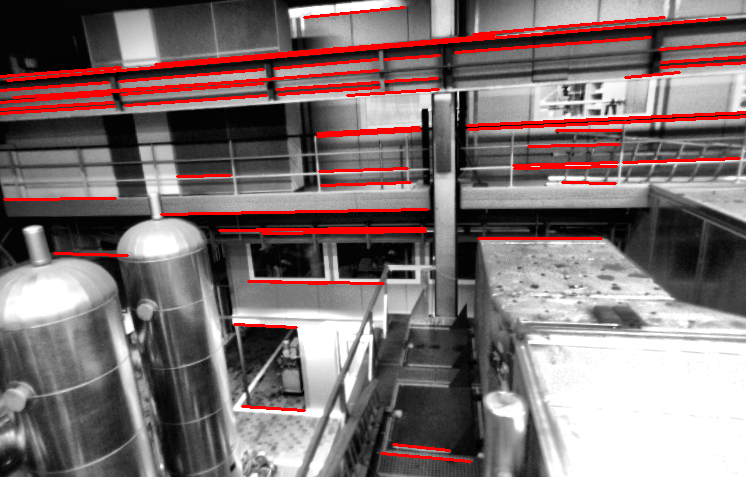}}
    \end{minipage}\hfill
    \begin{minipage}[c]{0.46\textwidth}    
    \subfigure[]{\label{fig:mapping:b}\includegraphics[width=\linewidth]{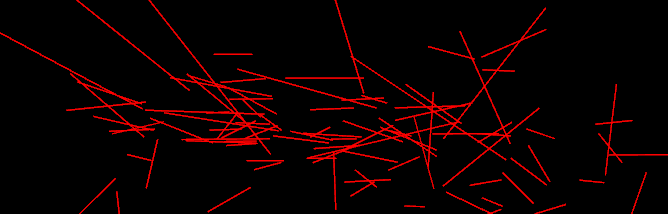}}
    \subfigure[]{\label{fig:mapping:c}\includegraphics[width=\linewidth]{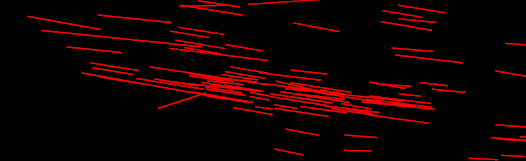}}
    \end{minipage}
    \caption{Line mapping result of VINS with a structural constraint. \subref{fig:mapping:a} Image with degenerate lines represented in red for $\textit{machine hall}$ in EuRoC dataset. Parallel lines on the wall cause degeneracy during triangulation. \subref{fig:mapping:b} Top view of line mapping result without structural constraint. \subref{fig:mapping:c} Top view of line mapping result with structural constraint.}
    \label{fig:line_mapping}
\end{figure*}

\begin{figure*}[ht!]
    \centering
    \includegraphics[width=\linewidth]{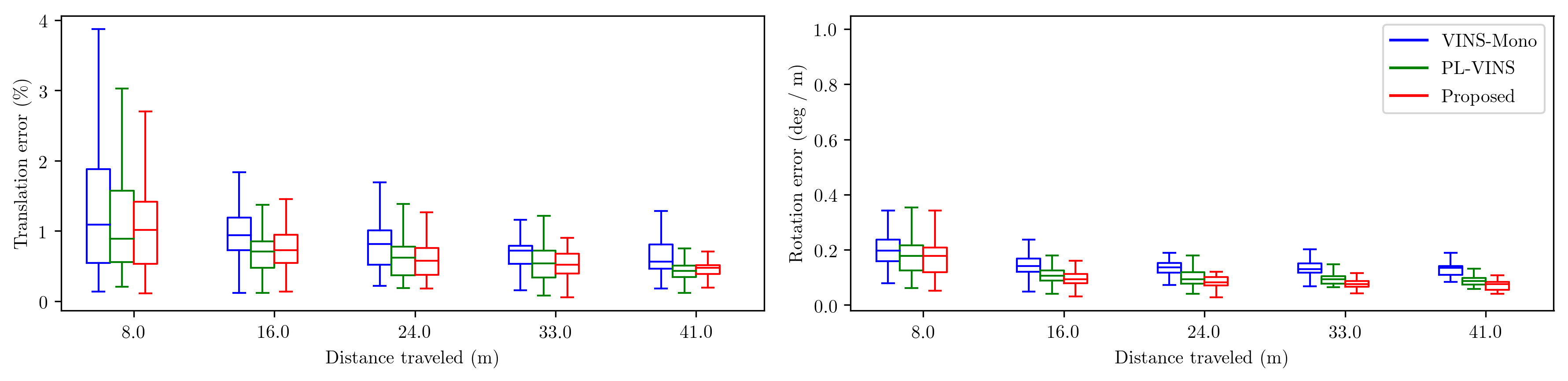}
    \caption{Boxplots of RMSE according to distance traveled for VINS-Mono, PL-VINS, and the proposed algorithm, respectively, for \textit{V2\_02\_medium} in EuRoC dataset.}
    \label{fig:boxplot}
    % \vspace{-5mm}
\end{figure*}

\begin{figure}[ht!]
    \label{fig:V2_02_trajectory_top}
    \includegraphics[width=\linewidth]{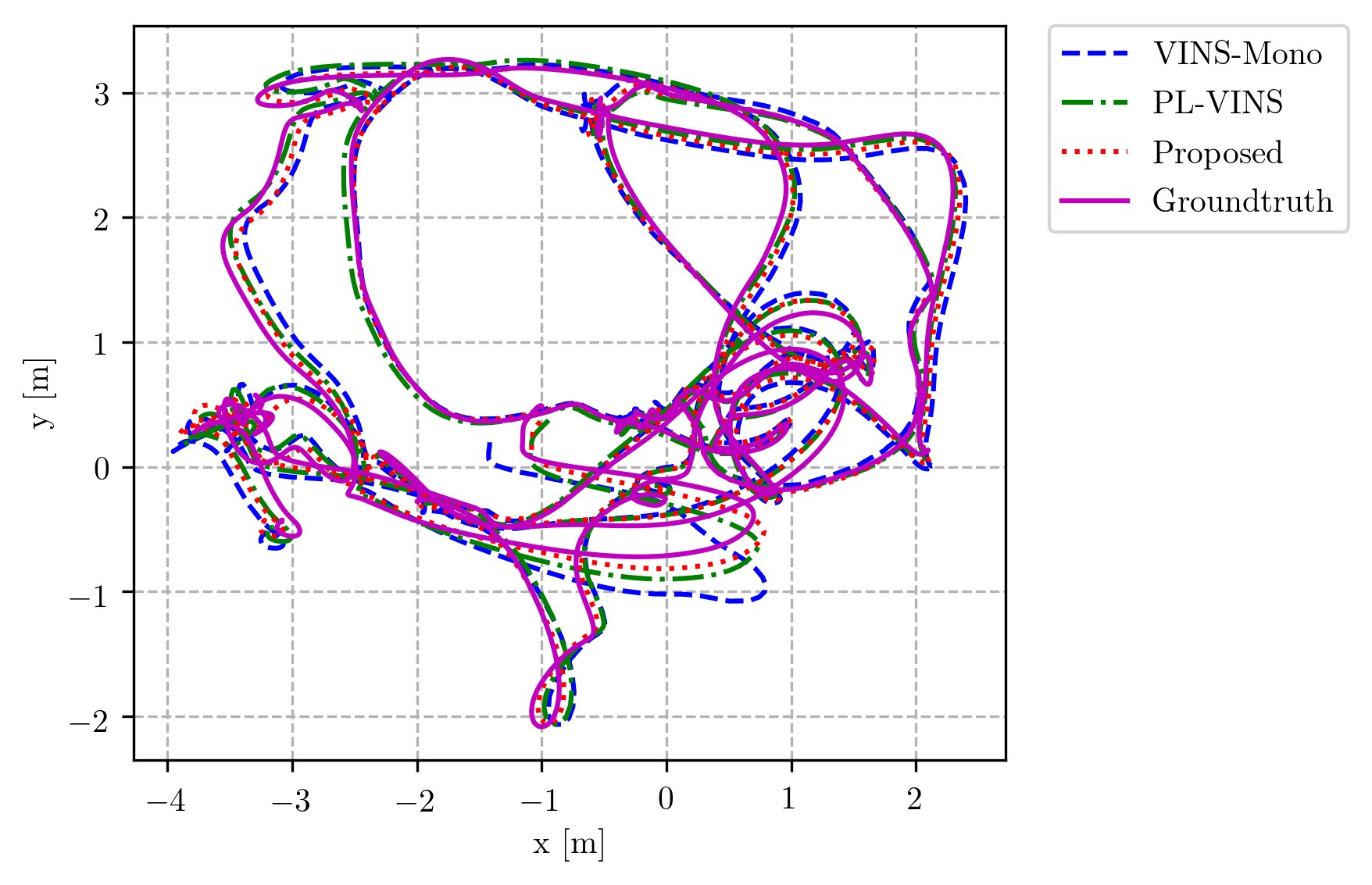}
    % \subfigure[]{\label{fig:V2_02_trajectory_side}\includegraphics[width=\linewidth]{figure/V2_02_trajectory_side.png}}
    \caption{Top view of trajectory estimated with VINS-Mono, PL-VINS, and the proposed algorithm for \textit{V2\_02\_medium} in EuRoC dataset.}
    % \subref{fig:V2_02_trajectory_top}~Trajectory
    % \subref{fig:V2_02_trajectory_side}~Trajectory seen from the side.
    \label{fig:traj}
    % \vspace{-5mm}
\end{figure}

For accurate algorithm evaluation, we used the rpg trajectory evaluation tool suggested by Zhang~\textit{et al.} \cite{zhang2018tutorial} and the results are shown in Figs.~{\ref{fig:boxplot}} and \ref{fig:traj}. TABLE~\ref{table:error} shows the translational root mean square error (RMSE) and maximum error for the EuRoC dataset. 

\begin{table}[h]
\caption{Translational RMSE and maximum error for the EuRoC Dataset (Unit: m)}
\label{table:error}
\begin{tabular}{*{7}{c}}\toprule
\multirow{3}{*}{} &\multicolumn{2}{c}{VINS-Mono} & \multicolumn{2}{c}{PL-VINS} & \multicolumn{2}{c}{Proposed} \\ 
Translation       & RMSE & Max. & RMSE & Max. & RMSE & Max. \\ \midrule
MH\_01\_easy      & 0.164 & 0.383 & 0.283 & 0.582 & \textbf{0.142} & \textbf{0.270} \\
MH\_02\_easy      & 0.140 & 0.381 & 0.195 & 0.546 & \textbf{0.126} & \textbf{0.346} \\
MH\_03\_medium    & 0.225 & 0.575 & \textbf{0.187} & \textbf{0.377} & 0.198 & 0.453 \\
MH\_04\_difficult & 0.408 & 0.610 & 0.335 & 0.588 & \textbf{0.301} & \textbf{0.448} \\
MH\_05\_difficult & 0.312 & 0.512 & 0.347 & 0.560 & \textbf{0.293} & \textbf{0.481} \\
V1\_01\_easy      & 0.094 & 0.209 & \textbf{0.071} & \textbf{0.150} & 0.087 & 0.200 \\
V1\_02\_medium    & 0.115 & 0.416 & 0.086 & 0.232 & \textbf{0.072} & \textbf{0.137} \\
V1\_03\_difficult & 0.203 & 0.401 & \textbf{0.152} & \textbf{0.348} & 0.156 & 0.375 \\
V2\_01\_easy      & 0.099 & \textbf{0.236} & \textbf{0.095} & 0.251 & 0.098 & 0.245 \\
V2\_02\_medium    & 0.161 & 0.569 & 0.120 & 0.287 & \textbf{0.103} & \textbf{0.211} \\
V2\_03\_difficult & 0.341 & 0.748 & 0.278 & \textbf{0.505} & \textbf{0.277} & 0.571 \\
\bottomrule 
\end{tabular}
% \vspace{-5mm}
\end{table}

% Overall, the proposed algorithm has about 15\% smaller errors than VINS-Mono. In addition, PL-VINS and the proposed algorithm showed similar performance in the \textit{easy} dataset with slow MAV motion, sufficient features, and little illumination change. However, the proposed algorithm is more accurate for \textit{medium} and \textit{difficult} datasets with low texture and variant illumination scenes. 
PL-VINS and the proposed algorithm have better performance than VINS-Mono overall. In particular, PL-VINS shows about 6.3\% smaller error than the proposed algorithm in four datasets out of 11 datasets. On the contrary, the proposed algorithm shows 25\% smaller error than PL-VINS in the remaining 7 datasets. In the proposed algorithm, more accurate results can be obtained because degenerate lines can be corrected through structural constraints while PL-VINS does nothing with degenerate lines.
The state estimation rate of the proposed algorithm is about 20Hz, which is somewhat slower than that of VINS-Mono running at about 28Hz, but it is not a problem since the IMU is updated very fast.

As shown in Fig.~{\ref{fig:mapping:a}}, for the \textit{machine hall} in the EuRoC dataset, degeneracy occurred in all lines facing the $x$-axis of the camera motion. Figs.~{\ref{fig:mapping:b}} and \subref{fig:mapping:c} show the line mapping results of the proposed algorithm with and without structural constraint, respectively. Through structural constraints, the degenerate lines are optimized as parallel lines.

\section{Conclusion}\label{sec:conclusion}
In summary, we developed a novel SLAM that utilizes line features efficiently. For robust tracking, the optical-flow-based line tracking method was proposed in line matching. Finally, we analyzed and resolved the degeneracy problem that must be considered when employing line features in SLAM. Experiments were performed with the EuRoC MAV dataset, from easy to difficult situations, to prove the validity of the proposed algorithm. The proposed method has better localization accuracy than VINS-Mono and PL-VINS. As a result, line tracking performance was increased by about 20\%, and translation RMSE was decreased by about 15\% compared to the base algorithm. It was also possible to obtain a more accurate and visually discernable line map by solving the degeneracy problem. In future work, we will apply an adaptive algorithm that finds the proper optimization factors between point and line features in SLAM. 

% or


% \section*{Acknowledgment}

% \ifCLASSOPTIONcaptionsoff
%   \newpage
% \fi


\bibliographystyle{IEEEtran}
\bibliography{reference}


%\end{IEEEbiography}


%\newpage



%\vfill


\end{document}